# Optimization of Flip-Landing Trajectories for Starship based on a Deep Learned Simulator

Liwei Chen[1*], Tong Qin[1,2*], Zhenhua Huangfu[2], Li Li[1], Wei Wei[1]

[1]Beijing Institute of Astronautical Systems Engineering, Beijing 100076, China
[2]Harbin Institute of Technology, Harbin, China
*Authors to whom correspondence should be addressed (e-mail):
jilinchl@163.com; qinshaomou@sina.com;

**Abstract:** We propose a differentiable optimization framework for flip-and-landing trajectory design of reusable spacecraft, exemplified by the Starship vehicle. A deep neural network surrogate, trained on high-fidelity CFD data, predicts aerodynamic forces and moments, and is tightly coupled with a differentiable rigid-body dynamics solver. This enables end-to-end gradient-based trajectory optimization without linearization or convex relaxation. The framework handles actuator limits and terminal landing constraints, producing physically consistent, optimized control sequences. Both standard automatic differentiation and Neural ODEs are applied to support long-horizon rollouts. Results demonstrate the framework's effectiveness in modeling and optimizing complex maneuvers with high nonlinearities. This work lays the groundwork for future extensions involving unsteady aerodynamics, plume interactions, and intelligent guidance design.

## 1. INTRODUCTION

Reusable launch vehicles represent a critical milestone in the advancement of next-generation space transportation systems. Among them, SpaceX's "Super Heavy-Starship" system stands as a remarkable exemplar, whose success has catalyzed transformative progress across the aerospace industry. A particularly challenging component of this system is the flip-and-landing maneuver, which involves transitioning from a high-angle belly-flop descent to a vertical orientation, followed by a soft landing under tight kinematic and control constraints.

This process exhibits strong nonlinearities and tight coupling across multiple physical domains, including aerodynamics, guidance, thrust vectoring, and attitude dynamics[1][2][3]. Conventional guidance and control designs often rely on simplified dynamics models, or resort to convex optimization methods that require linearizations or constraint relaxations[4][5][6]. While these approaches offer computational tractability, they typically compromise physical fidelity and may fail to capture critical interactions in high-angle or transitional regimes.

In this study, we propose a differentiable physics-aware framework that tightly couples a deep neural network-based aerodynamic model with a differentiable rigid-body dynamics solver. This enables end-to-end gradient-based optimization of flip-and-landing trajectories with full physical consistency, without the need for convex relaxation, model linearization, or system decoupling.

Our approach replaces traditional CFD or empirical drag models with a learned surrogate trained on high-fidelity Reynolds-Averaged Navier-Stokes (RANS) simulation data. The learned model is naturally differentiable and tightly integrated into the trajectory optimization loop. By leveraging this coupling, we preserve nonlinear dynamics fidelity, respect actuator constraints, and ensure convergence to physically valid solutions under realistic boundary and control conditions.

Key contributions of this work include:

1) Development of a differentiable trajectory optimization framework integrating deep-learned aerodynamics and rigid-body flight dynamics;
2) Demonstration of flip-and-landing trajectory optimization under actuator and terminal constraints, without convexification or simplification;
3) Exploration of Neural ODE-based optimization to support memory-efficient gradient propagation across long rollout horizons;
4) A modular structure allowing extension toward more advanced scenarios such as plume-aerodynamic interaction, unsteady flow, and disturbance-aware landing.

We believe this work provides a foundation for future multiphysics-aware intelligent aerospace design, where data-driven models and physics-based simulation are tightly integrated for complex maneuver planning and control..

## 2. RELATED WORK

### 2.1 Vertical-recovery Landing

The Starship employs a vertical landing strategy: during its terminal phase, it maintains a near-90° angle of attack in a belly-flop aerodynamic configuration, followed by multi-engine ignition and coupled thrust vectoring-aerodynamic control to achieve rapid attitude transition and sub-meter precision soft landing. This process involves large-scale attitude maneuvers and strongly nonlinear constraints, requiring coordinated resolution of multidimensional challenges such as aerodynamic-thrust coupling and dynamic constraint conflicts. Recent advances in convex optimization have demonstrated significant efficacy in aerospace trajectory optimization, particularly for real-time nonlinear systems. For instance, Açıkmeşe et al. [4] proposed lossless convexification for Martian landing, transforming non-convex thrust constraints into second-order cone programming (SOCP), improving computational efficiency by two orders of magnitude. Subsequent studies by Zhang et al. [5] and Lu et al [6] further validated convex optimization methods for rocket vertical landing and Starship flip maneuvers.

Despite its computational merits, convex optimization introduces inherent trade-offs when applied to nonlinear systems. Simplified dynamics approximations and constraint relaxations (e.g., obstacle avoidance) risk oversimplifying high-dimensional nonlinearities or compromising feasibility [7]. Additionally, dimension inflation from convex reformulations and incompatibility with automatic differentiation frameworks hinder real-time deployment and end-to-end learning. These limitations motivate our avoidance of convex relaxations. Instead, we directly leverage differentiable programming to solve flight dynamics equations, tightly coupled with aerodynamics learned model for forward physical modeling. In the this paper, our approach preserves full model fidelity and constraint integrity while exploiting gradient back propagation for control optimization. To our knowledge, such methodology remains under-explored in rocket vertical recovery research.

### 2.2 Deep Learning Model

Recent advances in computational fluid dynamics have witnessed growing adoption of deep learning models (DLM) as surrogates for expensive numerical solvers. The differentiable neural network architecture enables long-term rollouts with well-preserved backpropagation gradients as well as essential characteristics for supporting gradient-based inverse problem research [8], including spacecraft dynamic control and integrated trajectory optimization. Amos et al. [9] proposed using Model Predictive Control (MPC) as a foundation for differentiable policy classes in reinforcement learning within continuous state and action spaces. Chen et al. [10] applied UNet neural networks to study the classic inverse problem in fluid mechanics—"the minimal drag configuration of a fixed-volume object under laminar flow"—obtaining optimized configurations for arbitrarily specified Reynolds numbers ranging from quasi-Stokes flow to laminar separation states. Subsequently, the Google DeepMind team demonstrated successful cases of inverse problem-solving using graph neural networks trained on large datasets, including two-dimensional flow diverters, three-dimensional flow control surfaces, and airfoil aerodynamic optimization [11]. The three-dimensional flow control case particularly highlighted the powerful capabilities of general-purpose deep learning models in addressing inverse problems.

In this study, as a preliminary show case of the coupling framework, we trained a learned model trained on high quality CFD datasets, such that the model predicts key aerodynamic characteristics at a given condition, namely the angle of attack. In the future, we will integrate more complex learned model such as flowfield neural operator. The neural network model is naturally differentiable and can be seamlessly integrated with the gradient-based optimization algorithm.

**2.3 Deep-Learned Model Predictive Control**

The present study fundamentally aligns with machine learning for inverse design, where data-driven models enable computationally efficient implementations of closed-loop control frameworks, particularly model predictive control [12][13][14]. Unlike conventional reduced-order approaches (e.g., POD, DMD) [12], our methodology employs a nonlinear surrogate model for the aerodynamic force and moment prediction. Crucially, the model are tightly coupled with flight dynamics equations to perform full-order physical trajectory simulations, preserving the inherent nonlinearity of the system without linearization, a distinctive feature rarely explored in aerospace GNC research.

## 3. FLIPPING AND LANDING DYNAMICS

In this study, we investigate a coupled ordinary differential equation (ODE) and partial differential equation (PDE) system involving rigid body-fluid interaction through boundary conditions and control variables in a two-dimensional plane.

**3.1 Rigid Body Dynamics**

Considering the non-dimensional form, the reference length is taken as the Starship vehicle length $L_{\text{ref}}$=50m, the reference velocity as the sea-level speed of sound 335.57 m/s, and the reference mass as 24,000 kg. All other physical quantities are derived from these reference parameters using dimensional analysis to obtain their non-dimensional forms. The governing equations are as follows:

$$\dot{\mathbf{r}} = \mathbf{v}$$
$$\dot{\mathbf{v}} = (\mathbf{F}_T + \mathbf{F}_A \cdot \varepsilon)/m + \mathbf{g}$$
$$\dot{\theta} = \omega$$
$$\dot{\omega} = (M_T + M_A \cdot \eta)/J_z$$

$$\dot{m} = -1/I_{sp}T$$

$$\dot{\delta}_d = (\delta - \delta_d)/T_d$$

Here, $\mathbf{r} \in R^2$ denotes the aircraft's vector position in the 2D plane; $\mathbf{v} \in R^2$ represents the velocity vector; m is the aircraft mass; $F_T$ is the engine thrust vector; $M_T$ is the moment generated by the engine; $F_A$ denotes the fluid dynamic force acting on the aircraft; $M_A$ is the fluid-induced moment on the aircraft surface; g is the gravitational acceleration vector; θ is the pitch angle; $J_z$ is the moment of inertia about the center of mass; $I_{sp}$ is the engine-specific impulse; δ is the engine deflection angle. Following the recommendations in Reference [6], the engine deflection dynamics are modeled with $\delta_d$ as the response to δ, and $T_d$ as the dimensionless time constant of a first-order inertial element. Parameter values are provided in Table 1.

This approach integrates the coupling of flow fields with flight dynamics control, while operating within current practical design priorities and resource allocation frameworks. The aerodynamic and fluid dynamic analyses were conducted under the assumption of no engine exhaust jets — a simplification aligned with initial design-phase computational efficiency requirements. To address the inherent physical discrepancies between this assumption and real-world jet effects (where exhaust flows alter global pressure distributions and consequently influence aerodynamic forces/moments), we implemented angle-of-attack-dependent correction coefficients ε and η, derived from empirical validation studies. In this paper, no jet corrections are applied to aerodynamic forces and moments, i.e., both ε and η are set to 1.0.

### 3.2 Fluid Dynamics and Learned Model

The computational intensity of traditional Reynolds-Averaged Navier-Stokes (RANS) simulations presents significant challenges for optimal control tasks requiring coupled fluid/rigid-body integration. To address this limitation, we first constructed a comprehensive dataset spanning the operational envelope of interest.

The Mach number during the phase of interest is below 0.3, so we assume the flow is incompressible. The configuration under consideration is the Starship S31 prototype, capturing its characteristic state during flip-landing: forward flaps deployed flat while aft flaps are folded upward at 80° as depicted in Figure 1. We conduct RANS simulation with boundary layer resolved grids (~O(4M) grid points) using Fluent, and obtained the aerodynamic characteristics at AoA from 0 to 350 degree (one case every 10 degrees). Figure 1 shows the pressure distribution on the surface at 50° angle of attack.

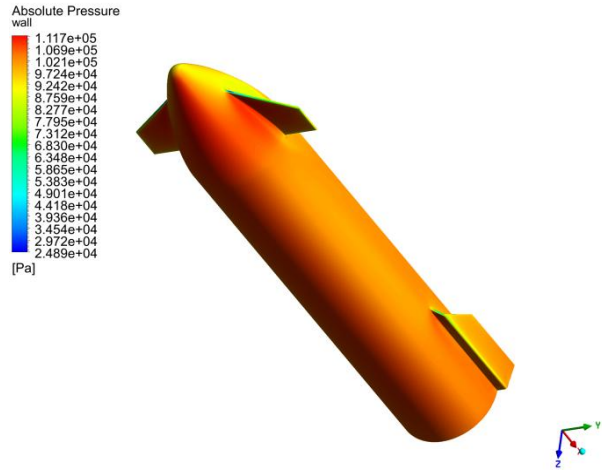

Figure 1. Surface pressure distribution on the Starship-like vehicle at an angle of attack of 50°, computed using RANS with boundary-layer resolved mesh. This snapshot is representative of the high-angle-of-attack regime encountered during the flip maneuver.

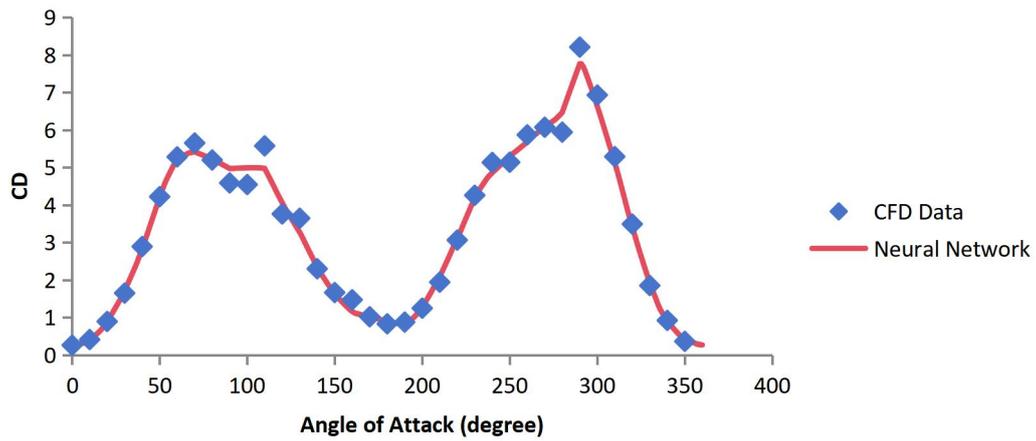

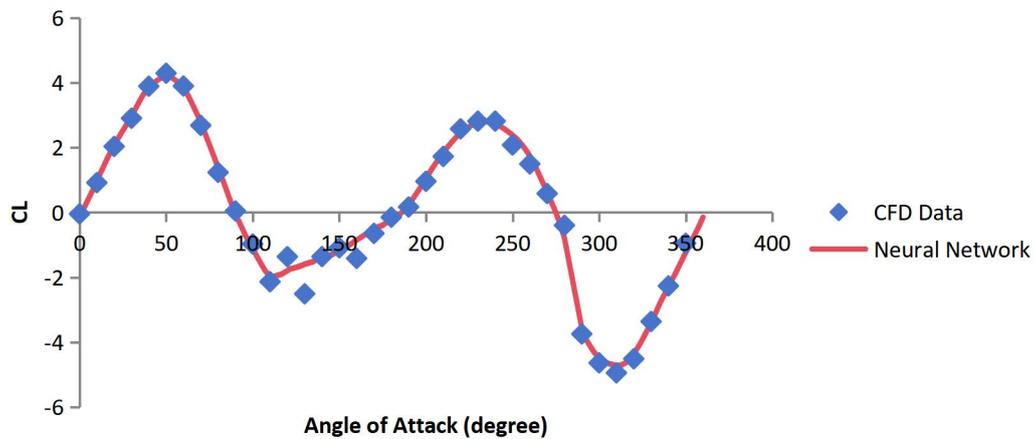

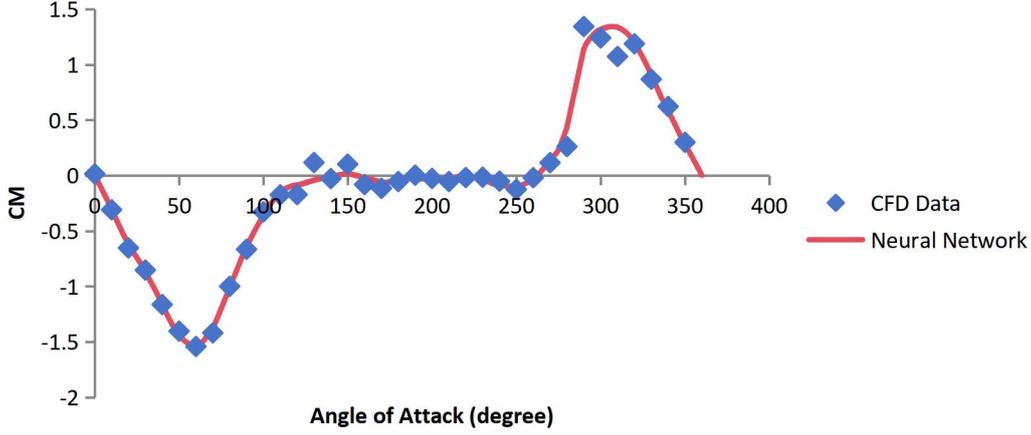

Figure 2. Comparison of aerodynamic coefficients predicted by the deep-learned model (DLM) and high-fidelity CFD simulations across a range of angles of attack. The DLM achieves excellent agreement, validating its accuracy for use in optimization.

Subsequently, we trained a neural network using MLP (denoted as $f_{NN}$) based on the dataset. The input to the network has two channels: the sine and cosine functions of the angle of attack ($\alpha_\infty$) in the present study, considering the periodicity. The angle of attack can be obtained from the rigid body dynamics:

$$\alpha_\infty = \arctan(v_y/v_x) - \theta.$$

The network outputs the aerodynamic characteristics:

$$[C_L, C_D, C_M] = f_{NN}(\alpha_\infty).$$

**Error! Bookmark not defined.** demonstrates close agreement in surface pressure distributions between DLM predictions and CFD data at transonic conditions. Then, the aerodynamic forces and moment $\mathbf{A} = [F_{Ax}, F_{Ay}, M_A]^T$ can be calculated in a differentiable way, seamlessly integrated with the optimization loop. We perform the discretization in the time domain for the rigid body dynamics, and obtain the following form:

$$\mathbf{X}_{n+1} = R(\mathbf{X}_n, \mathbf{A}_n, \mathbf{C}_n).$$

Here, $\mathbf{X} = [x, y, \theta, u, v, \omega]^T$ is the state vector of the rigid body; $\mathbf{A} = [F_{Ax}, F_{Ay}, M_A]^T$ contains the aerodynamic forces and moment; $\mathbf{C} = [T, \delta]$ is the control vector. The time marching method is 4th-order Runge-Kutta method.

During the descent phase, the rocket must satisfy various constraints:

**Thrust Magnitude Constraint**:

If the designed guidance control law exceeds the rocket's operational capabilities, precise landing cannot be guaranteed. Thus, thrust magnitude constraints must be imposed during optimization:

$$25\% \leq T/T_{max} \leq 100\%$$

Note: Individual engines have a throttle adjustment range of **25% to 100%**.

**Engine Gimbal Angle Constraint**:

This study focuses on 2D planar flight control with gimbal angle limits: $-10 \leq \delta \leq 10°$

**Precision Landing Constraint**:

The core objective of vertical recovery is achieving pinpoint landing with zero terminal states:

$$\text{Position: } r(t_f)=0;$$

$$\text{Velocity: } v(t_f)=0;$$

$$\text{Rotating speed: } \omega(t_f)=0$$

**Control Force Smoothness Constraint**: additional constraint to ensure smoothness of control forces.

|  | Name of parameters | Notations [Unit] | Values |
|---|---|---|---|
| Vehicle configurations | Moment of inertia | $J_z$ [kg m^2] | $1.25 \times 10^7$ in Case 1; $3.2 \times 10^7$ in Case 2; |
|  | Specific impulse | I [s] | 350 |
|  | Wet weight | $m_{wet}$ [t] | 135 |
|  | Dry weight | $m_{dry}$ [t] | 120 |
|  | Center of mass (From the nose cone tip) | $l_{cg} / L_{ref}$ | 60% |
|  | Length of the vehicle | $L_{ref}$ [m] | 50 |
| Constraints | Maximum thrust | $T_{max}$ [kN] | 2300 |
|  | Adjustable thrust range | [%] | 25-100 |
|  | Maximum deflection angle | $\delta_{max}$ [deg] | 10 |
|  | Minimum deflection angle | $\delta_{min}$ [deg] | -10 |
|  | Maximum time for flipping | Flipping time [s] | 2.4 |
|  | Initial pitch angle | $\theta_0$ [deg] | 170 |
|  | Initial position | $r_0$ [m] | [0 0]' |
|  | Initial velocity | $v_0$ [m/s] | [-18.82 -106.73]' |
|  | Initial acceleration | $a_0$ (m/s2) | [0 0]' |
|  | Initial angular speed | $\omega_0$ (rad / s) | 0 |
|  | Final pitch angle | $\theta_f$ (deg) | 90 |
|  | Final position | $r_f$ (m) | 1. [-360 -1200]' in Case 1; 2. [-287.5 -750]' in Case 2. |
|  | Final velocity | $v_f$ (m/s) | [0 -0.1]' |
|  | Final angular speed | $\omega_f$ (rad/s) | 0 |

**Table 1** Parameters and Constraints in the Flip-Landing Problem.

## 3.3 Optimization Algorithm

The optimization loop involves multi-step unrolling for forecasting while accumulating gradient information. As shown in Figure 3(a), in an individual optimization step (or iteration), we simulate the rigid body dynamics coupled with the learned model for aerodynamic coefficients. After K time steps (K=90 in this paper), we compute the final loss function and subsequently backpropagate the gradients to update the control parameters, which are time sequences. That is one complete optimization step.

We employ the Adam optimizer [15] ($\beta_1$=0.9, $\beta_2$=0.999). The learning rate for the optimization uses a cosine annealing learning rate scheduler that decays from $1\times10^{-3}$ to $1\times10^{-5}$. As shown in Figure 3(b), in a typical optimization task, we optimize control parameter sequences for totally 5000 steps to achieve a convergent state.

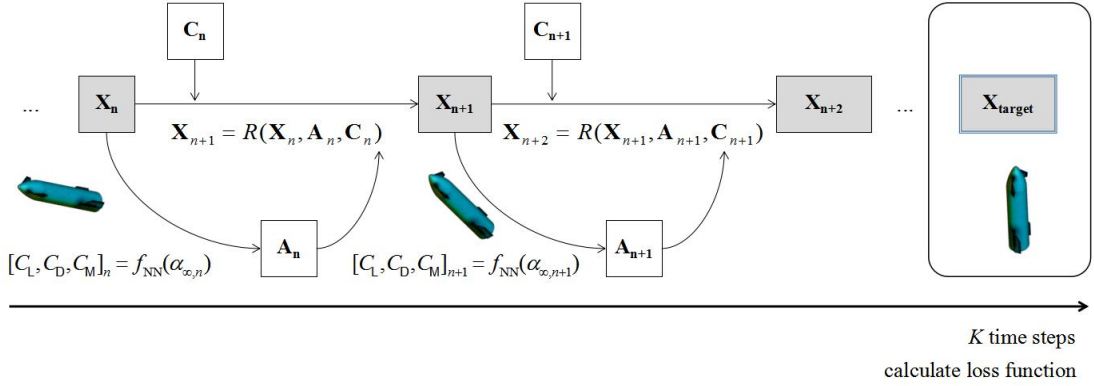

(a) Evaluation of loss function.

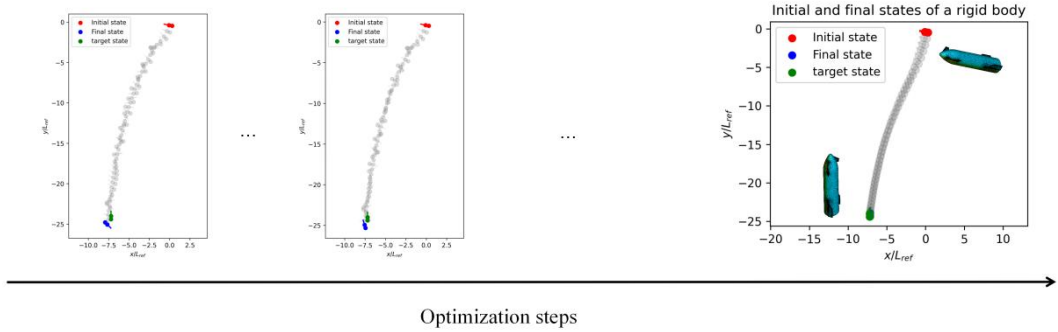

(b) A typical optimization loop.

Figure 3. (a) Schematic of a single optimization iteration: the coupled dynamics are rolled out over K steps, followed by loss evaluation. (b) Overall optimization loop using gradient backpropagation through time. The control sequence is updated iteratively to minimize trajectory error and satisfy constraints.

## 4. OPTIMIZATION RESULTS

**4.1 Case 1: with simplified Aerodynamics**

we adopted the simplified aerodynamic model from Reference [6], which neglects lift effects, assumes all aerodynamic forces are generated by an equivalent drag coefficient Cd, and maintains a fixed center of pressure position.

$$\mathbf{F}_A = \mathbf{D} + \mathbf{L} = -\tfrac{1}{2}\rho C_D \|\mathbf{v}\|\mathbf{v} S_{ref}$$

$$\mathbf{L} = 0$$

$$M_A = \mathbf{F}_A \times (\mathbf{x}_{cp} - \mathbf{x}_{cg}) = (l_{cp} - l_{cg})\tfrac{1}{2}\rho C_D S_{ref} \|\mathbf{v}\|(-\mathbf{v}) \times \begin{bmatrix} \cos\theta \\ \sin\theta \end{bmatrix}$$

Although this model has limitations, it is valuable for validating our algorithms before coupling with the learned model. To maintain consistency with Reference [6], we retained this simplified approach: fixed drag coefficients ($C_D$ = 0.5, 1.0, and 1.5 in three cases) and a fixed center of pressure at 0.55$L_{ref}$ (with the origin point defined at the rocket fairing tip, aligning with the literature).

Figure 4(a) illustrates the time histories of thrust magnitude, engine gimbal angle, horizontal velocity, and vertical velocity under the simplified aerodynamic model. The thrust and gimbal angle profiles reflect the coordinated control effort required to perform the rapid flip maneuver while respecting actuator constraints. Notably, the thrust increases sharply during the initial reorientation phase and gradually stabilizes for terminal descent, aligning with expectations for fuel-efficient braking and landing.

Figure 4(b) visualizes the attitude and trajectory evolution throughout the flip maneuver under the simplified aerodynamic model. The line segments represent vehicle orientation at each time step, and the points denote the center of mass and vehicle base. The vehicle starts from an initial vertical position y=0 and descends toward the landing site. The flip maneuver occurs at approximately y/$L_{ref}$=−6, corresponding to a descent depth of about 300 meters. After this point, the vehicle transitions into vertical descent with stable pitch control to meet the zero-terminal-state constraints. The optimized control inputs enable successful completion of the maneuver within the desired constraints and available thrust authority..

Figure 5 provides further insight into the dynamic state evolution. The effective torque from engine actuation initially spikes, enabling rapid reorientation, and then decreases as pitch stabilization is achieved. The vehicle's wet weight decreases smoothly due to propellant consumption. The angle of attack (AoA) follows a sharp trajectory consistent with the body flipping over, and the reduced frequency—a non-dimensional measure of unsteady aerodynamic response—peaks during the high-rate motion. These trends confirm that the optimization yields a physically plausible, constraint-satisfying solution for flip-landing under simplified aerodynamics.

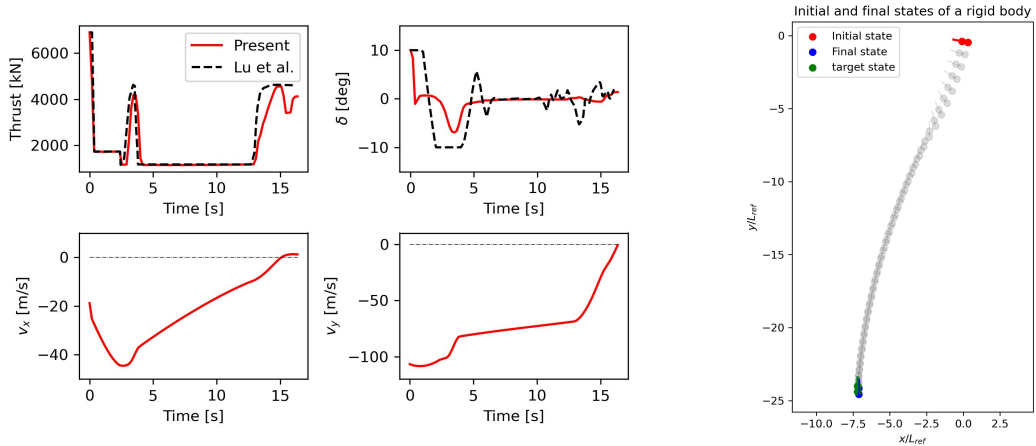

Figure 4. Case 1: Results using a simplified aerodynamic model. (a) Time history of thrust, engine gimbal angle, horizontal and vertical velocity. (b) Attitude and trajectory evolution during the flip maneuver.

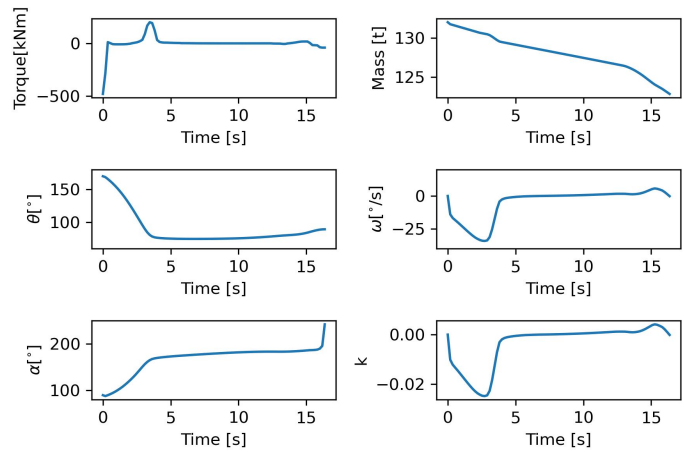

Figure 5. Case 1: Time evolution of key flight parameters after optimization. Shown are engine-generated torque, vehicle wet mass, pitch angle, angular velocity, angle of attack, and reduced frequency. These variables illustrate the dynamics of the flip and descent phases under simplified aerodynamics.

**4.2 Case 2: coupling with the DLM**

Figure 6(a) presents the optimized control histories with the deep-learned aerodynamic model (DLM) tightly coupled to the rigid body dynamics. Compared to Case 1, the thrust and gimbal angle profiles exhibit smoother transitions, highlighting the model's ability to exploit more nuanced aerodynamic effects captured by the DLM. The horizontal and vertical velocities show a controlled deceleration trend, ensuring precise touchdown while minimizing terminal velocity.

Figure 6(b) shows the trajectory and orientation evolution under the DLM-coupled dynamics. As in Case 1, the vehicle starts from an altitude of zero and performs a controlled descent. The flip maneuver occurs around y/$L_{ref}$=−8 (i.e., after descending approximately 400 meters). The smoother attitude evolution and more gradual pitch transition—compared to the simplified model—reflect the learned aerodynamic model's ability to provide more accurate force and moment predictions. This allows for improved coordination of control inputs, ultimately leading to a more stable terminal approach and precise landing.

In Figure 7, the evolution of key physical quantities further supports these observations. The pitch angle and angular velocity evolve more smoothly, indicating a better balance between aerodynamic torque and engine control. The refined behavior in these state variables confirms the added value of incorporating learned aerodynamics into the optimization loop.

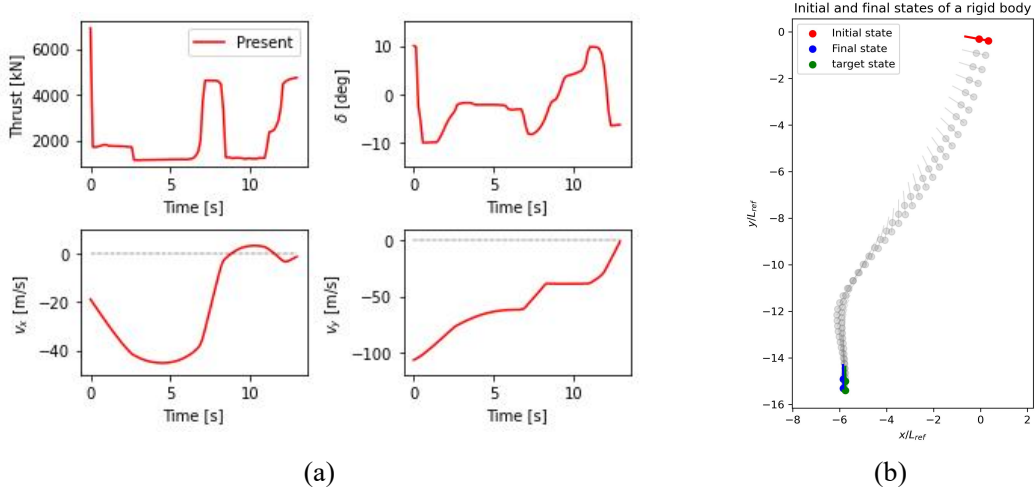

(a) (b)

Figure 6. Case 2: Results using deep-learned aerodynamic model (DLM). (a) Time history of control and velocity variables. (b) Attitude and trajectory evolution of the vehicle during descent.

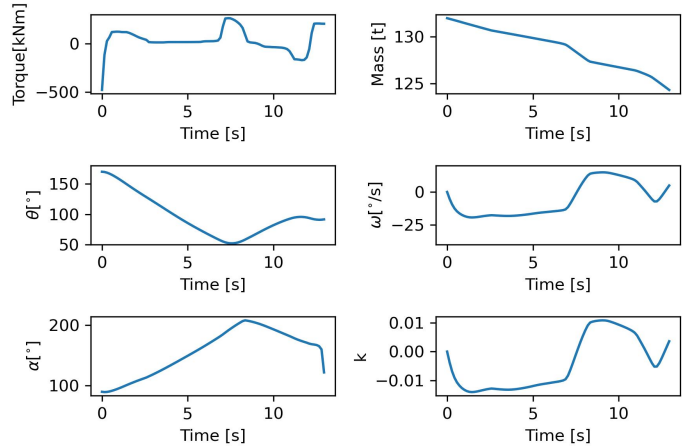

Figure 7. Case 2: Time evolution of key flight variables during flip and descent with DLM. Compared to Case 1, smoother transitions in pitch angle and angular velocity indicate improved control effectiveness from aerodynamic model fidelity.

### 4.3 Coupling with the Learned Model: Neural ODE for Case 2

In this section, we revisit Case 2 using the Neural ODE framework to mitigate memory bottlenecks during long rollout sequences. Although the control and trajectory results are not separately visualized, empirical comparisons confirm the consistency of the Neural ODE-based optimization with the vanilla automatic differentiation approach. Specifically, the mean absolute error (MAE) in control variables and state trajectories remains below 0.5%, indicating numerical equivalence.

More importantly, the memory efficiency of Neural ODEs enables scaling to longer horizon planning, which is particularly valuable in future scenarios involving extended reentry phases or disturbances. By enabling reversible gradient computation, the Neural ODE formulation preserves optimization quality while dramatically reducing computational overhead. This highlights its practical utility for large-scale trajectory planning.

## 5. CONCLUDING REMARKS

We integrated a learned model with three-degree-of-freedom (3-DoF) flight dynamics and implemented gradient-based optimal control for the Starship's flip-and-landing maneuver. Through comparative implementations of vanilla automatic differentiation and Neural ODE methodologies, we successfully derived physically consistent prediction estimates and control sequences, thereby demonstrating the framework's operational validity. Results indicate that the differentiable characteristics inherent in both rigid-body dynamics and neural network architectures enable robust gradient propagation across coupled physical domains, particularly when handling complex multi-constraint trajectories spanning hundreds of temporal steps.

This coupling framework establishes critical foundations for investigating advanced operational scenarios involving unsteady fluid-structure interactions and atmospheric disturbance analyses. By circumventing intricate convex optimization derivations, the methodology allows engineers to concentrate on forward process design while ensuring numerical stability. The proposed approach demonstrates significant potential for multidisciplinary co-design paradigms integrating trajectory planning, attitude control systems, and aerodynamic configurations.

We acknowledge limitations. First, the current deep neural network model operates as a static mapping rather than a neural operator capturing dynamic system evolution, neglecting unsteady aerodynamic effects during vehicle flipping. We also decoupled aerodynamics and propulsion effect, meaning that engine plume/external flow interactions are ignored. This is a pragmatic trade-off between high-fidelity coupled simulation costs and engineering applicability. Future work will develop high-fidelity databases incorporating unsteady aerodynamics, wind disturbances etc, followed by neural operator-based dynamic prediction models[17]. Such action-conditioned transition models will enable flexible planning/control and underpin digital twin-driven virtual flight platforms.

Second, the propagation of gradient information through extended computational chains constitutes a critical factor in optimization. While our experiments employ rollout lengths ranging from 90 to 180 steps, excessive rollout steps may induce numerical instability [18]. This challenge necessitates the implementation of advanced stabilization techniques [19][20], representing a crucial direction for our future research.

Third, Adam optimizer used for current scenarios is essentially first order, so landing problems with more complex constraints may require hybrid optimization or diffusion policy methods from robotics to overcome gradient-based limitations [21], which would be an interesting direction for future work.

## FUNDING

Funding for this research is provided by the Chinese National Science Foundation (grants

## AUTHOR CONTRIBUTIONS

Dr. Liwei Chen: Funding acquisition (PI); Conceptualization; Investigation (lead); Methodology (lead); Neural network design (lead); Validation (lead); Visualization (lead); Writing-original draft (lead); Writing-review & editing (lead); Project administration (lead); Supervision(lead).

Dr. Tong Qin: Investigation; Conceptualization; Writing-review & editing; Project administration.

Zhenhua Huangfu: CFD simulation; Training dataset generation.

Li Li: Writing-review & editing.

Wei Wei: Writing-review & editing; Project administration.